\documentclass[nohyperref]{article}

\usepackage{microtype}
\usepackage{graphicx}
\usepackage{subfigure}
\usepackage{booktabs} 
\usepackage{hyperref}

\usepackage[accepted]{icml2022}

\usepackage{amsmath}
\usepackage{amssymb}
\usepackage{mathtools}
\usepackage{amsthm}

\usepackage{amsmath,amsfonts,bm}

\def\eqref#1{equation~\ref{#1}}

\def\1{\bm{1}}

\def\vu{{\bm{u}}}
\def\vv{{\bm{v}}}
\def\vw{{\bm{w}}}
\def\vx{{\bm{x}}}

\def\vz{{\bm{z}}}

\DeclareMathAlphabet{\mathsfit}{\encodingdefault}{\sfdefault}{m}{sl}
\SetMathAlphabet{\mathsfit}{bold}{\encodingdefault}{\sfdefault}{bx}{n}

\def\gG{{\mathcal{G}}}

\def\gK{{\mathcal{K}}}

\def\sR{{\mathbb{R}}}

\newcommand{\Var}{\mathrm{Var}}

\DeclareMathOperator*{\argmin}{arg\,min}

\renewcommand{\epsilon}{\varepsilon}

\newcommand{\bj}{\mathbf{j}}

\newcommand{\by}{\mathbf{y}}

\newcommand{\bA}{\mathbf{A}}
\newcommand{\bB}{\mathbf{B}}
\newcommand{\bC}{\mathbf{C}}
\newcommand{\bD}{\mathbf{D}}
\newcommand{\bE}{\mathbf{E}}

\newcommand{\bG}{\mathbf{G}}

\newcommand{\bI}{\mathbf{I}}

\newcommand{\bK}{\mathbf{K}}
\newcommand{\bL}{\mathbf{L}}

\newcommand{\bU}{\mathbf{U}}

\newcommand{\bW}{\mathbf{W}}
\newcommand{\bX}{\mathbf{X}}
\newcommand{\bY}{\mathbf{Y}}
\newcommand{\bZ}{\mathbf{Z}}

\newcommand{\bOmega}{\bm{\Omega}}

\newcommand{\bTheta}{\bm{\Theta}}

\newcommand{\cE}{\mathcal{E}}

\newcommand{\cG}{\mathcal{G}}
\newcommand{\cH}{\mathcal{H}}
\newcommand{\cI}{\mathcal{I}}

\newcommand{\cM}{\mathcal{M}}

\newcommand{\cR}{\mathcal{R}}

\newcommand{\cT}{\mathcal{T}}

\newcommand{\cV}{\mathcal{V}}

\newcommand{\bbR}{\mathbb{R}}

\newcommand{\bone}{\mathbf{1}}
\newcommand{\norm}[1]{\ensuremath{\left\| #1 \right\|}}

\renewcommand{\vec}{\textup{\mbox{vec}}}

\usepackage{multirow}
\usepackage{enumerate}

\usepackage[capitalize,noabbrev]{cleveref}

\theoremstyle{plain}
\newtheorem{theorem}{Theorem}[section]
\newtheorem{proposition}[theorem]{Proposition}
\newtheorem{lemma}[theorem]{Lemma}

\theoremstyle{definition}

\newtheorem{assumption}[theorem]{Assumption}
\theoremstyle{remark}

\usepackage[textsize=tiny]{todonotes}

\begin{document}

\twocolumn[
\icmltitle{
Optimization-Induced Graph Implicit Nonlinear Diffusion
}



\icmlsetsymbol{equal}{*}

\begin{icmlauthorlist}
\icmlauthor{Qi Chen}{pku_math}
\icmlauthor{Yifei Wang}{pku_math}
\icmlauthor{Yisen Wang}{moe,pku_ai}
\icmlauthor{Jiansheng Yang}{pku_math}
\icmlauthor{Zhouchen Lin}{moe,pku_ai,pcl}
\end{icmlauthorlist}

\icmlaffiliation{pku_math}{School of Mathematical Sciences, Peking University, China}
\icmlaffiliation{pku_ai}{Institute for Artificial Intelligence, Peking University, China}
\icmlaffiliation{moe}{Key Lab. of Machine Perception (MoE), School of Artificial Intelligence, Peking University, China}
\icmlaffiliation{pcl}{Peng Cheng Laboratory, China}

\icmlcorrespondingauthor{Zhouchen Lin}{zlin@pku.edu.cn}

\icmlkeywords{Graph Neural Networks, Deep Equilibrium Models.}

\vskip 0.3in
]



\printAffiliationsAndNotice{}  

\begin{abstract}
Due to the over-smoothing issue, most existing graph neural networks can only capture limited dependencies with their inherently finite aggregation layers. To overcome this limitation, we propose a new kind of graph convolution, called Graph Implicit Nonlinear Diffusion (GIND), which implicitly has access to infinite hops of neighbors while adaptively aggregating features with nonlinear diffusion to prevent over-smoothing. Notably, we show that the learned representation can be formalized as the minimizer of an explicit convex optimization objective. With this property, we can theoretically characterize the equilibrium of our GIND from an optimization perspective. More interestingly, we can induce new structural variants by modifying the corresponding optimization objective. To be specific, we can embed prior properties to the equilibrium, as well as introducing skip connections to promote training stability. Extensive experiments show that GIND is good at capturing long-range dependencies, and performs well on both homophilic and heterophilic graphs with nonlinear diffusion. Moreover, we show that the optimization-induced variants of our models can boost the performance and improve training stability and efficiency as well. As a result, our GIND obtains significant improvements on both node-level and graph-level tasks. 
\end{abstract}

\section{Introduction}
In recent years, graph neural networks (GNNs) rise to be the state-of-the-art models for graph mining \cite{kipf2016semisupervised, velickovic2017graph, Hamilton2017inductive,li2022g2cn} and have extended applications in various scenarios such as biochemical structure discovery \cite{Gilmer2017neural, wan2019neodti}, recommender systems \cite{ying2018graph, fan2019graph}, natural language processing \cite{gao2019learning, Zhu2019graph}, and computer vision \cite{pang2017graph, valsesia2020deep}. 
Despite the success, these GNNs typically lack the ability to capture long-range dependencies. In particular, the common message-passing GNNs can only aggregate information from $T$-hop neighbors with $T$ propagation steps. 
However, existing works have observed that GNNs often degrade catastrophically when propagation steps $T\geq2$ \cite{li2018deeper}, a phenomenon widely characterized as over-smoothing. 
Several works try to alleviate it with more expressive aggregations of higher-order neighbors \cite{abu2019mixhop, zhu2020beyond}. Nevertheless, their ability to capture global information is inherently limited by the finite propagation steps.

Recently, implicit GNNs provide a new solution to this problem by replacing the deeply stacked explicit propagation layers with an implicit layer, which is equivalent to \emph{infinite} propagation steps \cite{gu2020implicit, liu2021eignn, park2021convergent}. Thereby, implicit GNNs could capture very long range dependencies and benefit from the global receptive field. Specifically, implicit GNNs achieve this by regularizing the explicit forward propagation to a root-finding problem with convergent equilibrium states. They further adopt implicit differentiation \cite{krantz2012implicit} directly through the equilibrium states to avoid long range back-propagation. As a result, the methods could get rid of performance degradation caused by explosive variance with more depth \cite{zhou2021understanding} while having constant memory footprint even with infinite propagation steps. These advantages indicate that implicit GNNs are promising alternatives to existing explicit GNNs.

The performance of implicit GNNs is largely determined by the design of the implicit layer. However, it is overlooked by previous works. Their implicit layers are direct adaptations of recurrent GNNs \cite{gu2020implicit} and lack theoretical justifications of their diffusion properties. Notably, a major drawback is that their aggregation mechanisms correspond to a \emph{linear isotropic diffusion} that treats all neighbors equally. However, as noted by recent theoretical discussions \cite{oono2019graph}, this isotropic property is exactly the cause of the over-smoothing issue. Thus, it is still hard for them to benefit from long range dependencies. This problem reveals that the infinite depth itself is inadequate. The design of the diffusion process, which decides the quality of the equilibrium states, is also the key to the actual performance of implicit GNNs. 

Motivated by this situation, in this work, we propose a novel and principled implicit graph diffusion layer, whose advantages are two folds. \textbf{First}, drawing inspirations from anisotropic diffusion process like PM diffusion \cite{perona1990scale}, we extend the linear isotropic diffusion to a more expressive \emph{nonlinear diffusion} mechanism, which learns nonlinear flux features between node pairs before aggregation.
In particular, our design of the nonlinear diffusion ensures that more information can be aggregated from similar neighbors and less from dissimilar neighbors, making node features less likely to over-smooth. \textbf{Second}, we can show for the first time that the equilibrium of our implicit nonlinear diffusion is the solution to a convex optimization objective. Based on this perspective, we can not only characterize theoretical properties of the equilibrium states, but also derive new structural variants in a principled way, \emph{e.g.}, adding regularization terms to the convex objective.

Based on the above analysis, we propose a model named Graph Implicit Nonlinear Diffusion (GIND). Several recent works have tried to connect GNN propagations to structural optimization objectives \cite{zhu2021interpreting,yang2021graph, ma2021unified, liu2021elastic}. However, their frameworks only consider aggregation steps and ignore the nonlinear transformation of features, which is also crucial in GNNs. In comparison, our GIND admits a unified objective of both the \emph{nonlinear} diffusion step and the transformation step. Therefore, our framework is more general, as it could take the interaction between diffusion and transformation into consideration. Last but not least, compared to previous optimization-based GNNs whose propagation rule is inspired by one single optimization step, our GIND directly models the equilibrium of the implicit layer as a minimizer of a convex objective (thus we call it optimization-induced). 
This shows that our GIND enjoys a much closer connection to the optimization objective compared to previous works. 

We evaluate GIND on a wide range of benchmark datasets, including both node-level and graph-level classification tasks. The results demonstrate that our GIND effectively captures long-range dependencies and outperforms both explicit and implicit GNN models in most cases. In particular, on heterophilic graphs, our nonlinear diffusion achieves significant improvements over previous implicit GNNs with isotropic diffusion. We further verify that two structural variants of GIND induced by principled feature regularization can indeed obtain consistent improvements over the vanilla GIND, which demonstrates the usefulness of our optimization framework.
In summary, our contributions are:
\begin{itemize}
    \item We develop a new kind of implicit GNNs, GIND, whose nonlinear diffusion overcomes the limitations of existing linear isotropic diffusion by adaptively aggregating nonlinear features from neighbors.
    \item We develop the first optimization framework for an implicit GNN by showing that the equilibrium states of GIND correspond to the solution of a convex objective. Based on this perspective, we derive three principled structural variants with empirical benefits.
    \item Extensive experiments on node-level and graph-level classification tasks show our GIND obtains state-of-the-art performance among implicit GNNs, and also compares favorably to explicit GNNs. 
\end{itemize}

\section{Preliminaries on Implicit GNNs}
Consider a general graph $\cG = (\cV, \cE)$ with node set $\cV$ and edge set $\cE$, where $|\cV| = n$ and $|\cE| = m$. Denote node features as $\vx \in \bbR^{n\times p}$, and the adjacency matrix as $\bA \in \bbR^{n\times n}$, where $\bA_{i,j} = 1$  if $(i,j) \in \cE$ and $\bA_{i,j} = 0$ otherwise. 
Denote the normalized adjacency matrix as $\hat{\tilde{\bA}} = \tilde{\bD}^{-1/2}\tilde{\bA} \tilde{\bD}^{-1/2}$. Here $\tilde{\bA}= \bA + \bI$ is the augmented adjacency matrix, and $\tilde\bD = \operatorname{diag}(d_1, \dots, d_n)$ is the degree matrix of $\tilde\bA$, where $d_j = \sum_j \tilde\bA_{i,j}$. 

Recently, several works  \cite{gu2020implicit,liu2021eignn,park2021convergent} have studied implicit GNNs. They are motivated by recurrent GCNs that employ the same transformation in each layer as follows: 
\begin{equation}
    \bZ_{k+1} = \sigma(\hat{\tilde{\bA}}\bZ_{k} \bW + b_{\bOmega}(\bX)), \quad k=1,2,\dots,\infty, 
\end{equation}
where $\bW$ is a weight matrix, $b_{\bOmega}(\bX)$ is the embedding of the input features $\bX$ through an affine transformation parameterized by $\bOmega$, and $\sigma(\cdot)$ is an element-wise activation function. 
In fact, it models the limiting process when the above recurrent iteration is applied for infinite times, 
\emph{i.e.}, $\bZ=\lim _{k \to \infty} {\bZ _k}$. 
As a result, the final equilibrium $\bZ$ potentially contains global information from all neighbors in the graph. 
Different from explicit GNNs,
the output features $\bZ$ of a general implicit GNN are directly modeled as the solution of the following equation, 
\begin{equation}
    \bZ = \sigma(\hat{\tilde{\bA}}\bZ\bW + b_{\bOmega}(\bX)). \label{eq:eq1} 
\end{equation}
The prediction is given by $\bY = g_{\bTheta} (\bZ)$, where $g_{\bTheta}$ is a trainable function parameterized by $\bTheta$. 

In practice, in the forward pass, we can directly find the equilibrium $\bZ$ via off-the-shelf black-box solvers like Broyden's method \cite{broyden1965class}. While in the backward pass, one can also 
analytically differentiate through the equilibrium by the implicit function theorem \cite{krantz2012implicit}, 
which does not require storing any intermediate activation values and uses only constant memory \cite{bai2019deep}. Several alternative strategies have also been proposed to improve its training stability \cite{geng2021on}.

\textbf{Limitations.}   
As shown above, existing implicit GNN models adopt the same feature propagation as the canonical GCN \cite{kipf2016semisupervised}. As discussed in many previous works \cite{oono2019graph}, this linear isotropic propagation matrix $\hat{\tilde{\bA}}$ will inevitably lead to feature over-smoothing after infinite propagation steps. Although this problem could be partly addressed by the implicit propagation process that admits a meaningful fixed point $\bZ$, the isotropic nature will still degrade the propagation process by introducing extra feature noises from dissimilar neighbors, as we elaborate below. 

\section{Graph Implicit Nonlinear Diffusion}
In view of the limitations of existing implicit GNNs based on linear isotropic diffusion, in this section, we propose a new nonlinear graph diffusion process for the design of implicit GNNs. Inspired by anisotropic diffusion, our nonlinear diffusion could adaptively aggregate more features from similar neighbors while separating dissimilar neighbors. Thus, its equilibrium states will preserve more discriminative features. 

\subsection{Nonlinear Diffusion Equation}

\textbf{Formulation.} Given a general continuous manifold $\cM$ where a feature function $\vu$ resides, along with operators such as the gradient and divergence that reside on the manifold $\cM$, one can model diffusion processes on $\cM$ \cite{eliasof2021pde, chamberlain2021grand}. 
In particular, we consider a \emph{nonlinear diffusion} at time $t$: 
\begin{equation}
    \partial_t \vu = \operatorname{div} (\gK^* \sigma (\gK \nabla \vu)), \label{eq:eq2}
\end{equation}
where $\gK$ is a linear operator, $\gK^*$ is its adjoint operator, $\nabla$ is the gradient operator, $\operatorname{div}$ is the divergence operator, and $\sigma(\cdot)$ is an element-wise transformation. 
It can be understood as a composition of two sequential operations, one is the nonlinear flux $\bj$ (describing differentials across a geometry) induced by the gradient $\nabla\vu$, \emph{i.e.},
\begin{equation}
 \bj = -\gK^* \sigma (\gK \nabla \vu),
\end{equation}
and the other is the continuity equation satisfying
\begin{equation}
    \partial_t \vu = - \operatorname{div}\bj.
\end{equation}

\textbf{Generalization of Isotropic Diffusion.}
One notable degenerated case of the nonlinear diffusion (\cref{eq:eq2}) is the linear isotropic diffusion, where we choose $\sigma$ and $\gK$ to be identity mappings as follows:
\begin{equation}
    \partial_t \vu = \operatorname{div} (\nabla \vu)=\Delta\vu. \label{eq:eq3}
\end{equation}
It admits a closed-form solution $\vu(t) =e^{t\Delta}\vu_0$ with an initial value  $\vu_0$ at $t=0$, where $\Delta$ is the Laplacian operator. \citet{wang2021dissecting} recently show that GCN propagation is equivalent to a time-discretized version of this equation. 
From this perspective, our nonlinear diffusion is a nontrivial generalization of isotropic diffusion in two ways: first, we add a linear operator $\gK$ inside the Laplacian operator for flexible parameterization \cite{nitzberg1992nonlinear} to adapt the diffusion to the initial value, and second, we introduce a nonlinear activation function $\sigma$ to model nonlinear flux, which greatly enhances its expressiveness as in neural networks. 

\begin{figure}[t]
\vskip 0.2in
\begin{center}
\centerline{\includegraphics[width=\columnwidth]{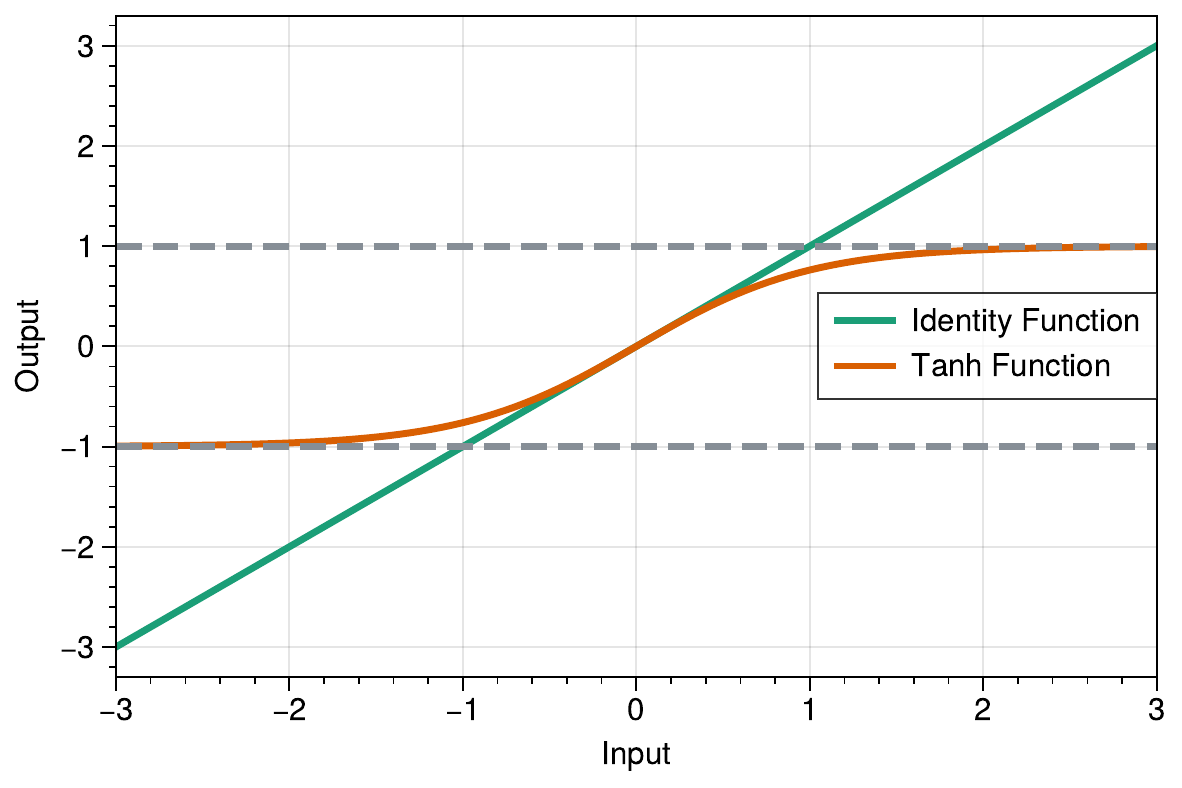}}
\caption{Comparison of two activation functions: $\sigma(x)=x$ and $\sigma(x)=\tanh(x)$. The nonlinear activation $\tanh(\cdot)$ keeps small values while shrinking large values.}
\label{fig:fig2}
\end{center}
\vskip -0.2in
\end{figure}

\textbf{Anisotropic Property.} Notably, this generalization enables us to incorporate anisotropic properties into the graph diffusion through specific choice of the activation function $\sigma(\cdot)$. To see this, ignoring $\gK$, we have $\bj=-\sigma(\nabla\vu)$ and the following relationship on their $\ell_2$-norms:
\begin{equation}
    \|\bj\|^2=\|\sigma(\nabla\vu)\|^2=\left(\int \left[\sigma(\nabla\vu(x))\right]^2dx\right)^{1/2}.
\end{equation}
The norm of the flux $\bj$ describes the magnitude of the information flow between two nodes. The larger the norm, the higher the degree of mixing between node pairs. 
In this case, comparing $\sigma(x)=x$ and $\sigma(x)=\tanh(x)$ as in \cref{fig:fig2}, we can see that the nonlinear activation $\tanh$ will keep small-value gradients while shrinking large-value gradients.
As a result, if the difference between two nodes is large, we think they are dissimilar, and then restrict the information exchange to prevent them from being undistinguished. 
Otherwise, if the difference is small, we tend to think they are similar, and let them exchange information like normal GCNs. 
In graph diffusion, this amounts to a desirable anisotropic-like behavior that adaptively aggregates more from similar neighbors and less from dissimilar neighbors, which helps prevent over-smoothing and improves robustness to noisy perturbations. 
Besides, our graph diffusion resembles the well-known PM diffusion \cite{perona1990scale}, which is a well-known adaptive diffusion used in image processing to preserve desired image structures, such as edges, while blurring others. 
To achieve this, they manually design a nonlinear function that approximates an impulse function close to edges to reweight the differences between image pixels. 
Compared with them, our nonlinear diffusion with the linear operator $\gK$ allows us to parameterize a flexible and learnable aggregation function.

\subsection{Proposed GIND}
The discussion above shows that our nonlinear diffusion is a principled generalization of previous linear isotropic diffusion with beneficial anisotropic properties. In this part, we apply it on the graph data and develop the corresponding implicit graph neural networks.

\textbf{Nonlinear Diffusion on Graphs.} 
As known in previous literature \cite{chung1997spectral}, the differential operators could be instantiated on a discrete graph $\gG$ with $n$ nodes and $m$ edges, and each node contains $h$-dimensional features.
Specifically, let $\bU\in\sR^{n\times h}$ be the feature matrix,
the gradient operator $\nabla$ corresponds to the incidence matrix $\bG\in\sR^{m\times n}$. It is defined as  $\bG_{k,i}=1$ if edge $k$ enters node $i$, $-1$ if it leaves node $i$, and $0$ otherwise. The divergence operator, the negative of which is the adjoint of the gradient operator, now corresponds to $-\bG^\top\in\sR^{n\times m}$. The linear operator $\gK$ corresponds to a feature transformation matrix $\bK\in\sR^{h\times h}$, and its adjoint $\gK^\star$ becomes $\bK^\top$. 
As a result, the nonlinear diffusion in \cref{eq:eq2} has the following matrix form on graph $\gG$,
\begin{equation}
    \partial_t\bU= -\bG^ \top\sigma(\bG\bU\bK^ \top)\bK. \label{eq:eq71}
\end{equation}
Following the analysis above, we choose the activation function $\sigma(x)=\tanh(x)$. A more detailed derivation can be found in \cref{sec:sigma}. 

\textbf{Our Implicit GNNs.}
By adopting the nonlinear graph diffusion developed above for the implicit graph diffusion mechanism, 
we develop a new implicit GNN, \textbf{G}raph \textbf{I}mplicit \textbf{N}onlinear \textbf{D}iffusion (\textbf{GIND}), with the following formulation,   
\begin{subequations}
    \begin{align}
    \bZ &= -\hat{\bG}^ \top\sigma(\hat{\bG}(\bZ + b_{\bOmega}(\bX))\bK^ \top)\bK, \label{eq:eq41} \\ 
    \hat{\bY} &= g_{\bTheta} (\bX+\bZ), \label{eq:eq42}
    \end{align}
\end{subequations}
where $\hat{\bG} = \bG \tilde{\bD}^{-1/2} /\sqrt{2}$ is the normalized incidence matrix. 
Here, we first embed the input feature matrix $\bX$ with an affine transformation $b_{\bOmega}(\cdot)$ with parameters $\bOmega$. 
Then, the input embedding is injected to the implicit diffusion layer, whose output $\bZ$ is the equilibrium of a nonlinear fixed point equation (\cref{eq:eq41}).
Afterwards, we use $\bX+\bZ$, the sum of the input features (initial value) and the equilibrium (the flux), as the final value to predict the labels. The readout head $g_{\bTheta}$ can be parametrized by a linear layer or an MLP. We also provide a row-normalized variant of the initial formulation (\cref{eq:eq41}) in \cref{sec:rownorm}. 

Notably, in GIND, we design the equilibrium states $\bZ$ to be the residual refinement of the input features $\bX$ through the diffusion process, \emph{a.k.a.~}the transported mass of $\bX$ \cite{weickert1998anisotropic}. As a result, starting from an initial value, the estimated transported mass $\bZ$ could be gradually refined through the fixed point iteration of our nonlinear diffusion process. Finally, it will reach a stable equilibrium $\bZ$ that cannot be further improved.

As an implicit model, our GIND enjoys the benefits of general implicit GNNs, including the ability to capture long-range dependencies as well as constant memory footprint. With our proposed nonlinear diffusion mechanism, it could adaptively aggregate useful features from similar neighbors and filter out noisy ones. Last but not least, as we show in the next section, the equilibrium states of our GIND can be formalized as the solution of a convex objective.
\section{An Optimization Framework for GIND}
In this section, inspired by recent works on optimization-based implicit models \cite{xie2021optimization}, we develop the first optimization framework for an implicit graph neural network. Specifically, we show that the equilibrium states of our GIND correspond to the solution of a convex objective. Based on this property, we show that we can derive principled variants of GIND through various regularization terms, which demonstrates a principled way to inject inductive bias into the model representations. 
\subsection{Formulation of Structural Objective}

\textbf{Notations.}
We use $\otimes$ to represent the Kronecker product, and use $\odot$ to represent element-wise product. 
For a matrix $\bW\in\sR^{p\times q}$, $\vw = \vec(\bW)$ represents the vectorized form of $\bW$ obtained by stacking its columns. We use $\norm{\cdot}_2$ for the matrix operator norm and $\norm{\cdot}$ for the vector $\ell_2$-norm. 
A function $f: \cH \to \bbR \cup\{+\infty\}$ is proper if the set $\{\vx: f(\vx) < +\infty\}$ is non-empty, where $\cH$ is the Euclidean space. For a proper convex function $f: \cH \to \bbR \cup\{+\infty\}$, its proximal operator $\text{Prox}^{\mu}_f(\vx)$ is defined as $\{ \vz \in \cH: \vz = \argmin _ {\vu} \frac{1}{2\mu}\norm{\vu-\vx}^2 + f(\vu)\}$. We omit $\mu$ when $\mu = 1$. 

For convenience, from now on, we adopt an equivalent “vectorized” version of the implicit layer (\cref{eq:eq41}) using Kronecker product:
\begin{align}
    \vz = -(\bK \otimes \hat{\bG})^ \top\sigma((\bK \otimes \hat{\bG})(\vz + \vec(b_{\bOmega}(\vx)))), \label{eq:eqimp} 
\end{align}
where $\vz$ is the vectorized version of the equilibrium state $\bZ$, and we use $f(\vz)$ to denote the right-hand side. 

The following theorem shows that the equilibrium states of
our GIND correspond to the solution of a convex objective.
\begin{theorem}
\label{thm:thm2}
Assume that the nonlinear function $\sigma(\cdot)$ is monotone and $L_\sigma$-Lipschitz, \emph{i.e.}, 
\begin{equation}
 0 \leq \frac{\sigma(a) - \sigma(b)}{a-b} \leq L_\sigma, \forall\ a,b \in \bbR, a \neq b,   
 \label{eqn:vector-implicit-layer}
\end{equation}
and  $1 \geq L_\sigma\norm{\bK \otimes \hat{\bG}}_2^2 = L_\sigma\norm{\bK}_2^2\norm{\hat{\bG}}_2^2$.
Then there exists a convex function $\varphi(\vz)$, such that its minimizer is the solution to the equilibrium equation $\vz = f(\vz)$. 
Furthermore, we have $\operatorname{Prox}_\varphi(\vz) = \frac{1}{L_\sigma+1}(L_\sigma\vz+f(\vz))$. 
\end{theorem}
With \cref{thm:thm2}, we can easily establish the following sufficient condition for our model to be well-posed, \emph{i.e.}, the existence and uniqueness of its solution, which guarantees the convergence of iterative solvers for reaching the equilibrium. 
\begin{proposition}
Our implicit layer (\cref{eq:eqimp}) is guaranteed to be well-posed if $\norm{\bK}_2 \|\hat{\bG}\|_2< 1$. Since we already have $\|\hat{\bG}\|_2 \leq 1$, we can fulfill the condition with $\norm{\bK}_2 < 1$. 
\label{pro:pro1}
\end{proposition}

\subsection{Optimization-Inspired Variants}
In previous part, we have established the structural objective of our proposed GIND. Here, we present an intriguing application of this perspective, which is to derive structural variants of our implicit model by injecting inductive bias in a principled way. Specifically, we study three examples: skip-connection induced by Moreau envelop \cite{xie2021optimization}, and two interpretable feature regularization: Laplacian smoothing and feature decorrelation.

\textbf{Optimization-Inspired Skip-Connection.} Since the equilibrium state of our implicit layer is the minimizer of an objective function $\varphi(\cdot)$, we can replace it directly with a new formula $\Phi(\cdot)$, as long as it has the same minimizer as the original one. Specifically, we choose its Moreau envelope function. 
Given a proper closed convex function $\varphi: \cH \to \bbR \cup\{+\infty\}$ and $\mu>0$, the Moreau envelope of $\varphi$ is the function
\begin{equation}
    M_{\varphi}^{\mu} (\vz) = \min_{\vu} \frac{1}{2\mu}\norm{\vu-\vz}^2 + \varphi(\vu),
\end{equation}
which is a smoothed approximate of $\varphi(\cdot)$ \cite{attouch1993approximation}. Meanwhile, the Moreau envelope function keeps the same critical points as the original one. 
Its proximal operator can be computed as follows:
\begin{equation}
\operatorname{Prox}^{\lambda}_{M_{\varphi}^{\mu}}(\vz) = \vz + \frac{\lambda}{\mu+\lambda} (\operatorname{Prox}^{\mu+\lambda}_{\varphi}(\vz) -\vz),
\end{equation}
where $\lambda, \mu >0$. 
Letting $\lambda=\alpha$ and $\mu=1-\alpha$, we have a new implicit layer induced from its proximal operator as follows: 
\begin{equation}
     \vz = \cT(\vz) \coloneqq (1-\alpha)\vz + \alpha f(\vz), \label{eq:eq6}
\end{equation}
where $\alpha \in (0, 1]$. 
Empirically, skip-connection improves stability in the iteration while keeping the equilibrium unchanged.

\textbf{Optimization-Inspired Feature Regularization.}
Regularization has become an important part of modern machine learning algorithms. Since we know the objective, we can combine it with regularizers to introduce customized properties to the equilibrium, 
which is equivalent to making a new implicit composite layer by appending one layer before the original implicit layer \cite{xie2021optimization}. Formally, if we modify $\Phi(\vz)$ to $\Phi(\vz) + \cR(\vz)$, then the implicit layer becomes:
\begin{equation}
    \vz = \cT(\vz) \circ \cT_{\cR},
\end{equation}
where $\cT_{\cR} = \operatorname{Prox}_{\cR}$, and $\circ$ denotes the mapping composition. 
When the proximal is hard to calculate, we can use a gradient descent step as an approximate evaluation, which is $\cT_{\cR_z} \approx \cI - \eta \frac{\partial\cR_z}{\partial z}$. In general, we can instantiate $\cR_z$ as any convex function that has the preferred properties of the equilibrium. Specifically, we consider two kinds of regularization. 
\begin{itemize}
\item \textbf{Laplacian Regularization.}
The graph Laplacian operator $\bL = \bD - \bA$ is a positive semi-definite matrix \cite{chung1997spectral}. We use the (symmetric normalized) Laplacian regularization $\hat{\tilde{\bL}} = \bI - \hat{\tilde{\bA}}$ to push the equilibrium of the linked nodes closer in the feature space. It is defined as follows:
\begin{equation}
    \cR_{\operatorname{Lap}}(\vz) = \vz^ \top \tilde{\bD}^{-\frac{1}{2}}L\tilde{\bD}^{-\frac{1}{2}} \vz = \norm{\hat{\bG}\vz}^2.
\end{equation}

\item \textbf{Feature Decorrelation.}
We use the feature decorrelation regularization to reduce redundant information between feature dimensions \cite{rodriguez2017regularizing, ayinde2019regularizing}. It is defined as follows:
\begin{equation}
    \cR_{\operatorname{Dec}}(\vz) = \frac{1}{2}\norm{\hat{\vz}\hat{\vz}^ \top-\bI}_F^2, 
\end{equation}
where $\hat{\vz}$ is the normalized $\vz$. 
\end{itemize}

\section{Efficient Training of GIND}
Training stability has been a widely existing issue for general implicit models \cite{bai2019deep,geng2021on,li2021optimization} as well as implicit GNNs \cite{gu2020implicit}. To train our GIND, we adopt an efficient training strategy that works effectively in our experiments.

\textbf{Forward and Backward Computation.} Specifically, in the forward pass, we adopt the fixed point iteration as in \citet{gu2020implicit}; while in the backward pass, we adopt the recently developed Phantom Gradient \cite{geng2021on} estimation,
which enjoys both efficient computation and stable training dynamics.

\textbf{Variance Normalization.} In previous implicit models \cite{bai2019deep, bai2020multiscale}, LayerNorm \cite{ba2016layer} is often applied on the output features to enhance the training stability of implicit models. 
Specifically, for GIND, 
we apply normalization before the nonlinear activation $\sigma(\cdot)$ and drop the mean items to keep it a sign-preserving transformation:
\begin{equation}
    \operatorname{norm}(\vv) = \frac{\vv}{\sqrt{\Var(\vv)+\epsilon}} \odot \bm{\gamma},
\end{equation}
where $\bm{\gamma}>0$ denotes positive scaling parameters and $\epsilon$ is a small constant. As discussed in previous works \cite{zhou2021understanding}, this could effectively prevent the variance inflammation issue and help stabilize the training process with increasing depth.

\begin{table*}[t]
\caption{Results on heterophilic node classification datasets: mean accuracy (\%) $\pm$ standard deviation over different data splits.}
\label{tab:tab1}
\vskip 0.15in
\begin{center}
\begin{tabular}{llccccc}
\toprule
Type      & Method   & Cornell        & Texas          & Wisconsin      & Chameleon      & Squirrel       \\\midrule
          & GCN      & 59.19$\pm$3.51 & 64.05$\pm$5.28 & 61.17$\pm$4.71 & 42.34$\pm$2.77 & 29.0$\pm$1.10  \\
          & GAT      & 59.46$\pm$6.94 & 61.62$\pm$5.77 & 60.78$\pm$8.27 & 46.03$\pm$2.51 & 30.51$\pm$1.28 \\
          & JKNet    & 58.18$\pm$3.87 & 63.78$\pm$6.30 & 60.98$\pm$2.97 & 44.45$\pm$3.17 & 30.83$\pm$1.65 \\
Explicit  & APPNP    & 63.78$\pm$5.43 & 64.32$\pm$7.03 & 61.57$\pm$3.31 & 43.85$\pm$2.43 & 30.67$\pm$1.06 \\
          & Geom-GCN & 60.81          & 67.57          & 64.12          & 60.9           & 38.14          \\
          & GCNII    & 76.75$\pm$5.95 & 73.51$\pm$9.95 & 78.82$\pm$5.74 & 48.59$\pm$1.88 & 32.20$\pm$1.06 \\
          & H2GCN    & 82.22$\pm$5.67 & 84.76$\pm$5.57 & 85.88$\pm$4.58 & 60.30$\pm$2.31 & 40.75$\pm$1.44 \\ \midrule
          & IGNN     & 61.35$\pm$4.84 & 58.37$\pm$5.82 & 53.53$\pm$6.49 & 41.38$\pm$2.53 & 24.99$\pm$2.11 \\
Implicit  & EIGNN    & 85.13$\pm$5.57 & 84.60$\pm$5.41 & 86.86$\pm$5.54 & 62.92$\pm$1.59 & 46.37$\pm$1.39 \\\cmidrule{2-7}
          & \textbf{GIND} (ours)& \textbf{85.68$\pm$3.83} & \textbf{86.22$\pm$5.19} & \textbf{88.04$\pm$3.97} & \textbf{66.82$\pm$2.37} & \textbf{56.71$\pm$2.07} \\ \bottomrule
\end{tabular}
\end{center}
\vskip -0.1in
\end{table*}

\section{Comparison to Related Work}
Here, we highlight the contributions of our proposed GIND through a detailed comparison to related work, including implicit GNNs and explicit GNNs.

\subsection{Comparison to Implicit GNNs} 

\textbf{Diffusion Process.}
Prior to our work, IGNN \cite{gu2020implicit} and EIGNN \cite{liu2021eignn} adopt the same aggregation mechanism as GCN \cite{kipf2016semisupervised}, which is equivalent to an isotropic linear diffusion that is irrelevant to neighbor features \cite{wang2021dissecting}. CGS \cite{park2021convergent} instead uses a learnable aggregation matrix to replace the original normalized adjacency matrix. However, the aggregation matrix is fixed in the implicit layer, thus the diffusion process still cannot adapt to the iteratively updated hidden states. 
In comparison, we design a nonlinear diffusion process with anisotropic properties. As a result, it is adaptive to the updated features and helps prevent over-smoothing. An additional advantage of GIND is that we can deduce its equilibrium states from a convex objective in a principled way, while in previous works the implicit layers are usually heuristically designed.

\textbf{Training.} 
IGNN \cite{gu2020implicit} adopts projected gradient descent to limit the parameters to a restricted space at each optimization step, which, still occasionally experiences diverged iterations. EIGNN \cite{liu2021eignn} and CGS \cite{park2021convergent} resort to linear implicit layers and reparameterization tricks to derive a closed-form solution directly, however, at the cost of degraded expressive power compared to nonlinear layers. 
In our GIND, we instead enhance the model expressiveness with our proposed nonlinear diffusion. During training, we regularize the forward pass with variance normalization and adopt the Phantom Gradient \cite{geng2021on} to obtain a fast and robust gradient estimate.

\subsection{Comparison to Explicit GNNs}

\textbf{Diffusion-Inspired GNNs.}
Diffusion process has also been applied to design explicit GNNs. 
\citet{atwood2016diffusion} design multi-hop representations from a diffusion perspective. 
\citet{xhonneux2020continuous} address continuous message passing based on a linear diffusion PDE. \citet{wang2021dissecting} point out the equivalence between GCN propagation and the numerical discretization of an isotropic diffusion process, and achieve better performance by further decoupling the terminal time and the propagation steps. \citet{chamberlain2021grand} also explore different numerical discretization schemes. They develop models based on both linear and nonlinear diffusion equations. However, their nonlinear counterpart does not outperform the linear one. \citet{eliasof2021pde} resort to alternative dynamics, including a diffusion process and a hyperbolic process, and design two corresponding models. 
In comparison, our GIND replaces the \textit{explicit} diffusion discretization with an \textit{implicit} formulation. 
The implicit formulation admits an equilibrium that corresponds to infinite diffusion steps, through which
we enlarge the receptive field while being free from manual tuning of the terminal time and the step size of the diffusion equation.

\textbf{Optimization-Inspired GNNs.} Prior to our work, there have been many works \cite{zhu2021interpreting, yang2021graph, ma2021unified, liu2021elastic} that establish a connection between different linear propagation schemes and the graph signal denoising with Laplacian smoothing regularization. By modifying the objective function, \citet{zhu2021interpreting} introduce adjustable graph kernels with different low-pass and high-pass filtering capabilities, \citet{ma2021unified} introduce a model that regularizes each node with different regularization strength, and \citet{liu2021elastic} enhance the local smoothness adaptivity of GNNs by replacing the $\ell_2$ norm by $\ell_1$-based graph smoothing. 
\citet{yang2021graph} focus on the iterative algorithms used to solve the objective, and introduce a model inspired by unfolding optimization iterations of the objective function. 
Their discussions are all limited to explicit layers and ignore the (potentially nonlinear) feature transformation steps. In comparison, our GIND discusses implicit layers, and admits a unified objective of both the nonlinear diffusion step and the transformation step. Besides, optimization-inspired explicit GNNs usually model a single iteration step, while our implicit model ensures that the obtained equilibrium is exactly the solution to the corresponding objective.

\section{Experiments}

In this section, we conduct a comprehensive analysis on GIND and compare it against both implicit and explicit GNNs on various problems and datasets. We refer to \cref{sec:exp} for the details of the data statistics, network architectures and training details. 
We implement our GIND based on the PyTorch Geometric library \cite{Fey/Lenssen/2019}. Our code is available at \url{https://github.com/7qchen/GIND}.

\subsection{Performance on Node Classification}
\begin{table}[t]
\caption{Results on homophilic node classification datasets: mean accuracy (\%).}
\label{tab:tab2}
\vskip 0.15in
\begin{center}
\begin{tabular}{llccc}
\toprule
Type      & Method    & Cora  & CiteSeer & PubMed \\ \midrule
          & GCN       & 85.77 & 73.68    & 88.13  \\
          & GAT       & 86.37 & 74.32    & 87.62  \\
          & JKNet     & 85.25 & 75.85    & 88.94  \\
Explicit  & APPNP     & 87.87 & 76.53    & 89.40  \\
          & Geom-GCN  & 85.27 & \textbf{77.99}    & 90.05  \\
          & GCNII     & \textbf{88.49} & 77.08    & \textbf{89.57}  \\
          & H2GCN     & 87.87 & 77.11    & 89.49  \\ \midrule
          & IGNN*     & 85.80 & 75.24    & 87.66  \\
Implicit  & EIGNN*    & 85.89 & 75.31    & 87.92  \\\cmidrule{2-5}
          & \textbf{GIND} (ours)      & 88.25 & 76.81    & 89.22  \\ \bottomrule
\end{tabular}
\end{center}
\vskip -0.1in
\end{table}

\begin{table}[t]
\caption{Results of micro-averaged F$1$ score on PPI dataset.}
\label{tab:tab3}
\begin{center}
\begin{tabular}{llc}
\toprule
Type      & Method    & Micro-F1 \\ \midrule
          & GCN       & 59.2     \\
          & GAT       & 97.3     \\
Explicit  & GraphSAGE & 78.6     \\
          & JKNet     & 97.6     \\
          & GCNII     & \textbf{99.5}     \\ \midrule
          & IGNN      & 97.0     \\
Implicit  & EIGNN     & 98.0     \\ \cline{2-3}
          & \textbf{GIND} (ours) & 98.4     \\ \bottomrule
\end{tabular}
\end{center}
\vspace{-0.1in}
\end{table}

\begin{table*}[t]
\caption{Results of graph classification: mean accuracy (\%) $\pm$ standard deviation over 10 random data splits.}
\label{tab:tab4}
\vskip 0.15in
\begin{center}
\begin{tabular}{llccccc}
\toprule
Type      & Method       & MUTAG        & PTC          & COX2         & PROTEINS     & NCI1         \\ \midrule
          & WL           & 84.1$\pm$1.9 & 58.0$\pm$2.5 & 83.2$\pm$0.2 & 74.7$\pm$0.5 & \textbf{84.5$\pm$0.5} \\
          & DCNN         & 67.0         & 56.6         & -            & 61.3         & 62.6         \\
Explicit  & DGCNN        & 85.8         & 58.6         & -            & 75.5         & 74.4         \\
          & GIN          & \textbf{89.4$\pm$5.6} & 64.6$\pm$7.0 & -            & 76.2$\pm$2.8 & 82.7$\pm$1.7 \\
          & FDGNN        & 88.5$\pm$3.8 & 63.4$\pm$5.4 & 83.3$\pm$2.9 & 76.8$\pm$2.9 & 77.8$\pm$1.6 \\ \midrule
          & IGNN*        & 76.0$\pm$13.4& 60.5$\pm$6.4 & 79.7$\pm$3.4 & 76.5$\pm$3.4 & 73.5$\pm$1.9 \\
Implicit  & CGS          & \textbf{89.4$\pm$5.6} & 64.7$\pm$6.4 & -            & 76.3$\pm$4.9 & 77.6$\pm$2.0 \\
          & \textbf{GIND} (ours) & 89.3$\pm$7.4 & \textbf{66.9$\pm$6.6} & \textbf{84.8$\pm$4.2} & \textbf{77.2$\pm$2.9} & 78.8$\pm$1.7 \\ \bottomrule
\end{tabular}
\end{center}
\vskip -0.1in
\end{table*}
\textbf{Datasets.}
We test GIND against the selected set of baselines for node classification task.
We adopt the $5$ heterophilic datasets: Cornell, Texas, Wisconsin, Chameleon and Squirrel \cite{pei2019geom}. 
And for homophilic datasets, we adopt $3$ citation datasets: Cora, CiteSeer and PubMed. We also evaluate GIND on PPI to show that GIND is applicable to multi-label multi-graph inductive learning. 
For other datasets except PPI, we adopt the standard data split as \citet{pei2019geom} and report the average performance over the $10$ random splits. While for PPI, we follow the train/validation/test split used in GraphSAGE \cite{Hamilton2017inductive}. 

\textbf{Baselines.}
Here, we compare our approach against several representative explicit and implicit methods that also adopt the same data splits. 
For explicit models, we select several representative baselines to compare, \emph{i.e.}, GCN \cite{kipf2016semisupervised}, GAT \cite{velickovic2017graph}, JKNet \cite{xu2018representation}, APPNP \cite{klicpera2018predict}, Geom-GCN \cite{pei2019geom}, GCNII \cite{chen2020simple}, and H2GCN \cite{zhu2020beyond}. For Geom-GCN, we report the best result among its three model variants. 
For implicit methods, we present the results of IGNN \cite{gu2020implicit} and EIGNN \cite{liu2021eignn}. We implement IGNN on citation datasets with their $1$-layer model used for node classification, and implement EIGNN on citation datasets with their model used for heterophilic datasets. We mark the results implemented by us with $*$. We do not include CGS \cite{park2021convergent}, as they do not introduce a model for node-level tasks. 

\textbf{Results.} As shown in \cref{tab:tab1}, our model outperforms the explicit and implicit baselines on all heterophilic datasets by a significant margin, especially on the larger datasets Chameleon and Squirrel. In particular, we improve the current state-of-the-art accuracy of EIGNN from $46.37\%$ to $56.71\%$ on the Squirrel dataset, while the state-of-the-art explicit model H2GCN only reaches $40.75\%$ accuracy. 
Among explicit models, JKNet, APPNP and GCNII are either designed to consider a larger range of neighbors or to mitigate oversmoothing. Despite that they outperform GCN and GAT, they still perform worse than implicit models on most datasets. 
Compared to other implicit models (EIGNN and IGNN), GIND still shows clear advantages. Consequently, we argue that the success of our network stems not only from the implicit setting, but also from our nonlinear diffusion that enhances useful features in aggregation.

Also, our results in \cref{tab:tab2} show that GIND achieves similar accuracy to the compared methods on $3$ citation datasets, although they may not need as many long-range dependencies as the heterophilic datasets. 
On the PPI dataset, as depicted from \cref{tab:tab3}, our GIND achieves $98.4$ micro-averaged F$1$ score, still superior to other implicit methods and most explicit methods, close to the state-of-the-art models. 
Moreover, with a small initialization for the parameters, all our training processes are empirically stable. 

\subsection{Performance on Graph Classification}
\textbf{Datasets.}
For graph classification, we choose a total of $5$ bioinformatics benchmarks: MUTAG, PTC, COX2, NCI1 and PROTEINS \cite{yanardag2015deep}. Following identical settings as \citet{yanardag2015deep}, we conduct $10$-fold cross-validation with LIB-SVM \cite{chang2011libsvm} and report the average prediction accuracy and standard deviations in \cref{tab:tab4}. 

\textbf{Baselines.} Here, we include the baselines that also have reported results on the chosen datasets.
For explicit models, we choose Weisfeiler-Lehman Kernel (WL) \cite{shervashidze2011weisfeiler}, DCNN \cite{atwood2016diffusion}, DGCNN \cite{zhang2018end}, GIN \cite{xu2018representation} and FDGNN \cite{gallicchio2020fast}. 
For implicit models, we reproduce the result of IGNN \cite{gu2020implicit} with their source code for a fair comparison, since it used a different performance metric. We mark the results implemented by us with $*$. 
We do not include EIGNN \cite{liu2021eignn}, as they do not introduce a model for graph-level tasks. 

\textbf{Results.}
In this experiment, GIND achieves the best performance in $3$ out of $5$ experiments given the competitive baselines. Such performance further validates GIND’s success that it can still capture long-range dependencies when generalized to unseen testing graphs. Note that GIND also outperforms both implicit baselines. 

\subsection{Empirical Understandings of GIND}

\begin{table}[t]
\caption{Comparison of different regularization types.}
\label{tab:tab7}
\vskip 0.15in
\begin{center}
\begin{tabular}{lccc}
\toprule
Reg            & Cora                     & CiteSeer          & PubMed    \\ \midrule
None             & 88.25$\pm$1.25           & 76.81$\pm$1.68    & 89.22$\pm$0.40  \\
$\cR_{\operatorname{Lap}}$ & \textbf{88.33$\pm$1.15}  & \textbf{76.95$\pm$1.73} & \textbf{89.42$\pm$0.50}    \\
$\cR_{\operatorname{Dec}}$ & 88.29$\pm$0.92           & 76.84$\pm$1.70    & 89.28$\pm$0.41      \\
\bottomrule
\end{tabular}
\end{center}
\vskip -0.1in
\end{table}
\textbf{Feature Regularization.}
In this experiment, we compare the performance of two kinds of feature regularization: the Laplacian regularization $\cR_{\operatorname{Lap}}$ and the feature decorrelation $\cR_{\operatorname{Dec}}$. We use the $3$ citation dataset for comparison. We choose an appropriate regularization coefficient $\eta$ for each dataset. As reported in \cref{tab:tab7}, Laplacian regularization improves the performance for the homophilic datasets. We attribute this to the fact that similarity between nodes is an appropriate assumption for these datasets. The feature decorrelation also improves the performance slightly. 

\begin{figure}[t]
\vskip 0.2in
\begin{center}
\centerline{\includegraphics[width=\columnwidth]{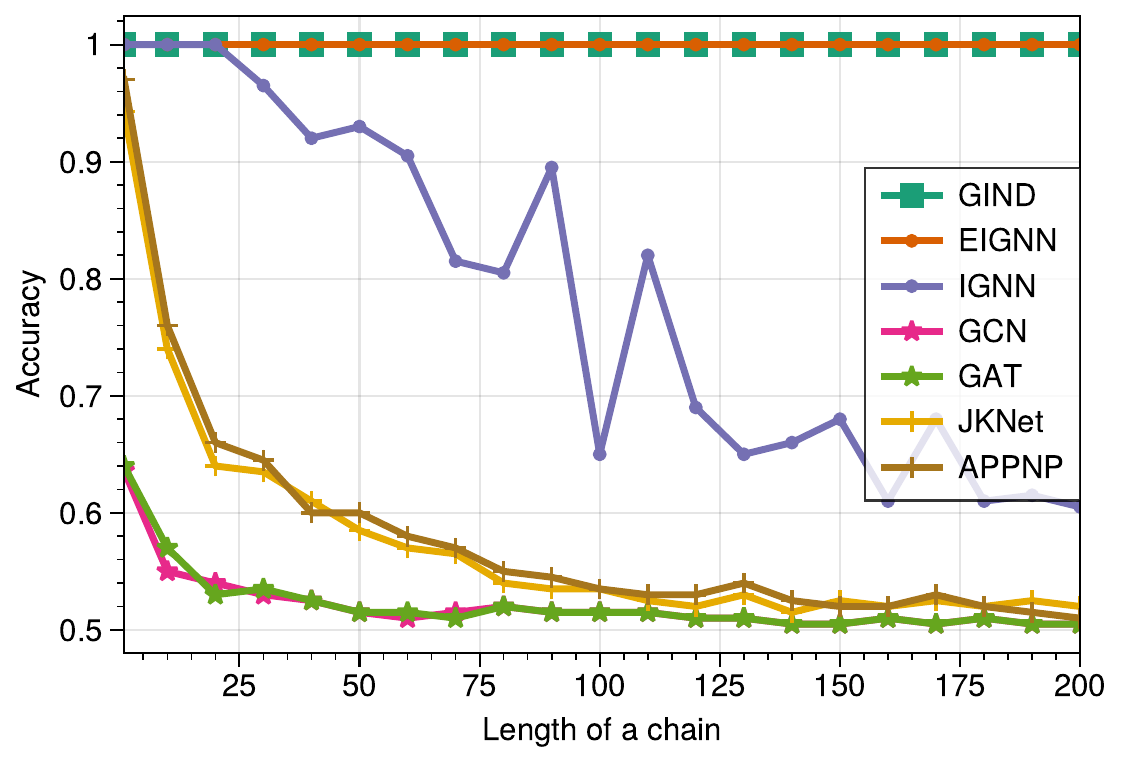}}
\caption{Average accuracy with respect to the length of chains.}
\label{fig:fig1}
\end{center}
\vspace{-0.2in}
\end{figure}

\textbf{Long-Range Dependencies.}
We follow \citet{gu2020implicit} and use the synthetic dataset Chains to evaluate models' abilities for capturing long-range dependencies. The goal is to classify nodes in a chain of length $l$, whose label information is only encoded in the starting end node. We use the same experimental settings as IGNN and EIGNN. 
We show in \cref{fig:fig1} the experimental results with chains of different lengths. 
In general, the implicit models all outperform the explicit models for longer chains, verifying that they have advantages for capturing long-range dependencies. 
GIND and EIGNN both repetitively maintain $100\%$ test accuracy with the length of $200$. While EIGNN only applies to linear cases, GIND can apply to more general nonlinear cases. 

\begin{table}[t]
\caption{Comparison of training time on the PubMed dataset. }
\label{tab:tab10}
\vskip 0.15in
\begin{center}
\begin{tabular}{lccc}
\toprule
Method      & Preprocessing & Training & Total        \\ \midrule
IGNN        & 0             & 83.67 s  & 83.67 s      \\
EIGNN       & 1404.45 s     & 69.14 s  & 1473.59 s    \\
CGS         & 0             & 118.65 s & 118.65 s     \\
GIND  (ours)      & \textbf{0}             & \textbf{47.31 s } & \textbf{47.31 s}      \\
\bottomrule
\end{tabular}
\end{center}
\vskip -0.1in
\end{table}

\textbf{Training Time.}
To investigate the training time, we train the $1$-layer models on PubMed dataset for $500$ epochs to compare their efficiency. 
Since we use gradient estimate for training, our model can be faster than those implicit models that compute exact gradients. 
As shown in \cref{tab:tab10}, our model costs much fewer time ($47.31$ s) than the three models that calculate exact gradients. Among the three models, EIGNN is the most efficient with $69.14$ s training cost, but its pre-processing costs much more than training. 

\begin{table}[t]
\caption{Comparison of linear and nonlinear diffusion in GIND.}
\label{tab:tab11}
\vskip 0.15in
\begin{center}
\begin{tabular}{lccc}
\toprule
Dataset     & CiteSeer         & Cornell         & Wisconsin      \\ \midrule
Linear      & 76.62$\pm$1.51   & 83.24$\pm$6.82  & 84.31$\pm$4.30 \\
Nonlinear   & \textbf{76.81$\pm$1.68} & \textbf{85.58$\pm$3.83} & \textbf{88.04$\pm$3.97} \\ \midrule\midrule
Linear      & 18.60s          & 19.94s         & 20.52s         \\ 
Nonlinear   & 20.44s          & 20.21s         & 21.30s         \\ \bottomrule
\end{tabular}
\end{center}
\vskip -0.1in
\end{table}
\textbf{Efficacy of Nonlinear Diffusion.}
We conduct an ablation of the proposed nonlinear diffusion by removing the nonlinearity in \cref{eq:eq41}. 
From \cref{tab:tab11}, we can see that the nonlinear diffusion has a clear advantage over linear ones, especially on heterophilic datasets. 
Meanwhile, the additional computation overhead brought by the nonlinear setting is almost neglectable.

\begin{figure}[t]
\vskip 0.2in
\begin{center}
\centerline{\includegraphics[width=\columnwidth]{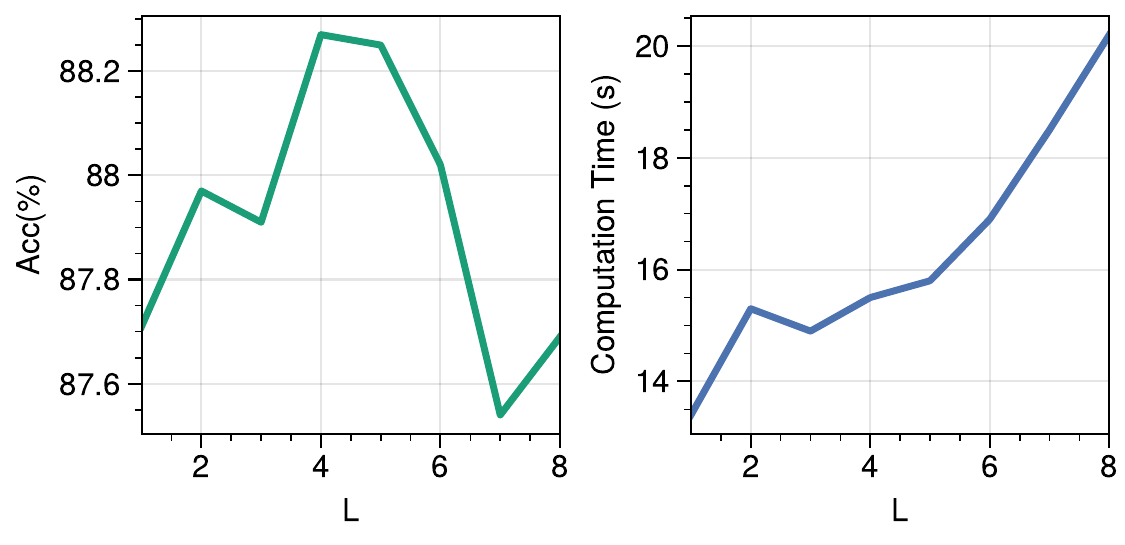}}
\caption{Left: test accuracy (\%) with increasing backpropagation steps $L$ on Cora. Right: the corresponding training time (s).}
\label{fig:fig3}
\end{center}
\vskip -0.2in
\end{figure}

\textbf{Gradient Estimate.}
In the left plot of \cref{fig:fig3}, we compare results with different backpropagation steps $L$ of Phantom Gradient \cite{geng2021on} along with their corresponding training time on the Cora dataset. We can see that there is a tradeoff between different steps, that either too large or too small an $L$ leads to degraded performance, and the sweet spot lies in $L=4$. The corresponding training time is $15.52$ s, which is still preferable to other implicit GNNs (Table \ref{tab:tab10}).

\section{Conclusion}
In this paper, we develop GIND, an optimization-induced implicit graph neural network, which has access to infinite hops of neighbors while adaptively aggregating features with nonlinear diffusion. 
We characterize the equilibrium of our implicit layer from an optimization perspective, and show that the learned representation can be formalized as the minimizer of an explicit convex optimization objective. 
Benefiting from this, we can embed prior properties to the equilibrium and introduce skip connections to promote training stability. 
Extensive experiments have shown that compared with previous implicit GNNs, our GIND obtains state-of-the-art performance on various benchmark datasets.

\section*{Acknowledgements}
Zhouchen Lin was supported by the major key project of PCL (grant No. PCL2021A12), the NSF China (No. 61731018), and Project 2020BD006 supported by PKU-Baidu Fund.
Yisen Wang was partially supported by the NSF China (No. 62006153) and Project 2020BD006 supported by PKU-Baidu Fund.

\bibliography{example_paper}
\bibliographystyle{icml2022}

\newpage
\appendix
\onecolumn

\section{Kronecker Product}
Given two matrices $\bA \in \bbR^{m\times n}$ and $\bB \in \bbR^{p\times q}$, the Kronecker product $\bA \otimes \bB \in \bbR^{pm \times qn}$ is defined as follows:
\begin{equation}
    \bA \otimes \bB = 
    \begin{pmatrix}
        \bA _{11}\bB & \cdots & \bA _{1n}\bB \\
        \vdots       & \ddots & \vdots \\
        \bA _{m1}\bB & \cdots & \bA _{mn}\bB
    \end{pmatrix}.
\end{equation}
By definition of the Kronecker product, we have the following important properties of the vectorization with the Kronecker product:
\begin{itemize}
    \item $\norm{\bA \otimes \bB} = \norm{\bA} \norm{\bB}$,
    \item $(\bA \otimes \bB) ^ \top = \bA ^ \top \otimes \bB ^ \top$,
    \item $\vec(\bA \bB \bC)  = (\bC ^ \top \otimes \bA)\vec(\bB)$. 
\end{itemize}

\section{Choice of the Nonlinear Transformation $\sigma(\cdot)$}\label{sec:sigma}
\textbf{Oriented Incidence Matrix on Undirected Graphs.}
If $\cG$ is undirected, we randomly assign an orientation for each edge to construct an oriented incidence matrix $\bG$, then it is unique to the obtained directed edge set $\tilde{\cE}$. Given $\bU \in \bbR^{n\times p}$ as the discrete version of $\vu$,
the gradient operator $(\bG\bU)_{k,\cdot} = \vu_j - \vu_i$ assigns the edge $k=(i,j) \in \tilde{\cE}$ the difference of its endpoint features. Similarly, the divergence operator $-\bG^\top$ assigns each node the sum of the features of all edges it shares. 

To ensure the discretization of the nonlinear diffusion is well-defined for undirected graphs, we need to ensure that randomly switching the direction of the edge would not influence the output. As a result, we make the following assumption.

\begin{assumption}
\label{ass:ass2}
The nonlinear function $\sigma(\cdot)$ is odd, \emph{i.e.}, $\sigma(-x) = -\sigma(x)$. 
\end{assumption}

Switching the direction of an edge $k=(i,j)$ is equivalent to multiplying a matrix $\bE_{k}$ to the left of $\bG$, where $\bE_k$ is obtained by switching the $(k,k)$-th element of an identity matrix to $-1$. If \cref{ass:ass2} holds, we have
\begin{align}
    \partial_t \bU &= -(\bE_{k}\bG)^ \top\sigma(\bE_{k}\bG\bU\bK^ \top)\bK, \\
    &= -\bG^ \top \bE_{k}^ \top\bE_{k}\sigma(\bG\bU\bK^ \top)\bK, \\ 
    &= -\bG^ \top \sigma(\bG\bU\bK^ \top)\bK.
\end{align}
Consequently, the discretization is well-defined for undirected graphs.

Besides \cref{ass:ass2}, 
as discussed in \cref{thm:thm2}, we need \cref{ass:ass1} to hold. Overall, $\tanh$ satisfies both assumptions, and has an effect to keep more small gradients while shrinking large gradients. 

\begin{assumption}
\label{ass:ass1}
The nonlinear function $\sigma(\cdot)$ is monotone and $L_\sigma$-Lipschitz, \emph{i.e.},
\begin{equation}
0 \leq \frac{\sigma(a) - \sigma(b)}{a-b} \leq L_\sigma, \forall a, b \in \bbR, a \neq b.
\end{equation}
\end{assumption}

\section{Proofs} \label{sec:proof}
\subsection{Conditions to be a Proximal Operator}
\begin{lemma} (modified version of Prop. 2 in \citet{gribonval2020characterization}). Consider $f: \cH \to \cH$ defined everywhere. The following properties are equivalent:
\begin{enumerate}[(i)]
    \item there is a proper convex \emph{l.s.c} function $\varphi: \cH \to \bbR \cup\{+\infty\}$ \emph{s.t. } $f(\vz) \in \operatorname{Prox}_\varphi (\vz)$ for each $\vz \in \cH$; \\
    \item the following conditions hold jointly:
        \begin{enumerate}
            \item there exists a convex \emph{l.s.c} function $\psi: \cH \to \bbR$ \emph{s.t. } $\forall \by \in \cH, f(\by) = \nabla \psi(\by)$;
            \item $\norm{f(\by) - f(\by ')} \leq \norm{\by - \by '}, \forall \by, \by' \in \cH$. 
        \end{enumerate}
\end{enumerate}
There exists a choice of $\varphi(\cdot)$ and $\psi(\cdot)$, satisfying (i) and (ii), such that $\varphi (\vz) = \psi ^*(\vz) - \frac{1}{2}\norm{\vz}^2$.

\begin{proof}
(i) $\Rightarrow$ (ii): Since $\varphi (\vx) + \frac{1}{2}\norm{\vx}^2$ is a proper \emph{l.s.c} $1$-strongly convex function, then by Thm. 5.26 in \citet{beck2017first}, its conjugate function $f^*(\by) = \sup \{\langle \by, \vx\rangle-f(\vx): \vx\in \cH\}$ is $\frac{1}{\sigma}$-smooth when $f$ is proper, closed and $\sigma$ strongly convex and vice versa. Thus, we have: 
\begin{equation}
    \psi (\vx) \coloneqq [\varphi(\vx) +  \frac{1}{2}\norm{\vx}^2]^*,
\end{equation}
is $1$-smooth with $\operatorname{dom}(\psi) = \cH$. Then we get:
\begin{align}
    f(\vx) &\in  \argmin _\vu  \frac{1}{2}\norm{\vu - \vx}^2 + \varphi (\vu) = \{\vu: \vx \in \partial \varphi(\vu) + \vu\}, \\
    &= \{\vu| \vx \in \partial (\varphi(\vu) + \frac{1}{2}\norm{\vu} ^2)\}, \\
    &= \{\vu| \vu = \nabla \psi(\vx)\} = \{\nabla \psi(\vx)\}.
\end{align}
Hence $f(\vx) = \nabla \psi(\vx)$, and $1$-smoothness of $\psi$ implies $f$ is nonexpansive. 

(ii) $\Rightarrow$ (i): Let $\varphi(\vx) = \psi ^* (\vx) - \frac{1}{2}\norm{\vx}^2$. 
Since $\psi(\vx)$ is $1$-smooth, similarly we can conclude: $\psi^*$ is $1$-strongly convex. Hence, $\varphi$ is convex, and:
\begin{align}
    \operatorname{Prox}_\varphi (\vx) &= \argmin _{\vu}\{ \frac{1}{2}\norm{\vu - \vx}^2 + \varphi (\vu)\}, \\
    &= \{\vu| \vx \in \partial \varphi(\vu) + \vu\}, \\
    &= \{\nabla \psi(\vx)\} = \{f(\vx)\},
\end{align}
which means $f(\vx) = \operatorname{Prox}_\varphi (\vx)$. 
\end{proof}
\label{lem:lem2}
\end{lemma}

\subsection{Proof of \cref{thm:thm2}}
The proof follows the one presented in \citet{xie2021optimization}. 
\begin{proof} 
The equation $\vz = f(\vz)$ can be reformulated as 
\begin{equation}
    \vz + L_\sigma\vz = L_\sigma\vz + f(\vz) \Longleftrightarrow \vz = g(\vz) \coloneqq \frac{1}{L_\sigma+1}(L_\sigma\vz+f(\vz)). 
\end{equation}
In the proof, without loss of generality, we let $L_\sigma = 1$. 
Since $\sigma(a)$ is a single-valued function with slope in $[0, 1]$, the element-wise defined operator $\sigma(a)$ is nonexpansive. Combining with $\norm{\bK \otimes \hat{\bG}} \leq 1$, operator $f(\vz)$ is nonexpansive, and $g(\vz)$ is also nonexpansive by definition. 

 Let $\tilde{\sigma}(a) = \int_0^a\sigma(t)dt$ be a function 
applied element-wisely to $\forall a \in \bbR$, and $\bone$ is the all one vector.
Since
$\bone^ \top \tilde{\sigma} (\by) = \sum _{i=1} ^n \tilde{\sigma} (y_i)$, we have $\nabla \tilde{\sigma} (\by) = [\sigma(y_1), \cdots, \sigma(y_n)]^ \top = \sigma (\by)$. 
Let $\psi(\vz) = \frac{1}{4}\norm{\vz}^2 - \frac{1}{2}\bone^ \top \tilde{\sigma}((\bK \otimes \hat{\bG})(\vz + \vec(b_{\bOmega}(\vx))))$, 
by the chain rule, $\nabla \psi(\vz) = \frac{1}{2}\vz - \frac{1}{2}(\bK \otimes \hat{\bG})^ \top\sigma((\bK \otimes \hat{\bG})(\vz + \vec(b_{\bOmega}(\vx)))) = g(\vz)$. 
Due to \cref{lem:lem2}, we have $g(\vz) = \operatorname{Prox}_\varphi (\vz)$, where $\varphi (\vz)$ can be chosen as $\psi^*(\vz) - \frac{1}{2}\norm{\vz}^2$. As a result, the solution to the equilibrium equation $\vz = f(\vz)$ is the minimizer of the convex function $\varphi(\cdot)$. 
\end{proof}

\section{Row-Normalization Variant of GIND} \label{sec:rownorm}
Considering numerical stability, we impose the symmetric normalization to the incidence matrix in our implicit layer (\cref{eq:eq41}), which is widely used in GNN models. Alternatively, we provide a row-normalization variant as follows. 
\begin{equation}
    \bZ = -(2\tilde{\bD})^{-1} \bG ^ \top\sigma(\bG(\bZ + b_{\bOmega}(\bX))\bK^ \top)\bK. \label{eq:eq16}
\end{equation}
Note that the two formulations are equivalent in the sense that 
\cref{eq:eq16} can be rewritten as the following formulation: 
\begin{equation}
    \bar{\bZ} = -\hat{\bG}^ \top\sigma(\hat{\bG}(\bar{\bZ} + \bar{b}_{\bOmega}(\bX))\bK^ \top)\bK,
\end{equation}
where $\bar{\bZ} = (2\tilde{\bD})^{\frac{1}{2}}\bZ$ and $\bar{b}_{\bOmega}(\bX) = (2\tilde{\bD})^{\frac{1}{2}}b_{\bOmega}(\bX)$. 
Moreover, without changing the output $\bY$, the row-normalized version of GIND has the following formulation: 
\begin{subequations}
    \begin{align}
    \bar{\bZ} &= -\hat{\bG}^ \top\sigma(\hat{\bG}(\bar{\bZ} + \bar{b}_{\bOmega}(\bX))\bK^ \top)\bK, \\
    \bY &= g_{\bTheta} (\bX+ (2\tilde{\bD})^{-\frac{1}{2}}\bar{\bZ}).
    \end{align}
\end{subequations}
Without loss of generality, all the results can be easily adapted to the row-normalization case. 
\section{More on Experiments} \label{sec:exp}
\subsection{Datasets}
The statistics for the datasets used in node-level tasks is listed in \cref{tab:tab9}. 
Among heterophilic datasets, Cornell, Texas and Wisconsin are web-page graphs of the corresponding universities, while Chameleon and Squirrel are web-page graphs of Wikipedia of the corresponding topic. The node features are bag-of-word representations of informative nouns in the web-pages. 
Among homophilic datasets, PPI contains multiple graphs where nodes are proteins and edges are interactions between proteins. 
The statistics for the datasets used in graph-level tasks is listed in \cref{tab:tab8}. All these datasets consist of chemical molecules where nodes refer to atoms while edges refer to atomic bonds. 
\begin{table*}[t]
\caption{Dataset statistics for node classification task.}
\label{tab:tab9}
\vskip 0.15in
\begin{center}
\begin{tabular}{lcccccccc}
\toprule
Dataset   & Cornell        & Texas          & Wisconsin      & Chameleon      & Squirrel   & Cora  & CiteSeer & PubMed   \\ \midrule
Nodes     & 1,283          & 183            & 251            & 2,277          & 5,201      & 2,708   & 3,327     & 19,717    \\
Edges     & 280            & 295            & 466            & 31,421         & 198,493    & 5429   & 4732     & 44338    \\
Features  & 1,703          & 1,703          & 1,703          & 2,325          & 2,089      & 1433   & 3703     & 500      \\
Classes   & 5              & 5              & 5              & 5              & 5          & 7      & 6        &   3        \\ 
\bottomrule
\end{tabular}
\end{center}
\vskip -0.1in
\end{table*}

\begin{table*}[t]
\caption{Dataset statistics for graph classification task.}
\label{tab:tab8}
\vskip 0.15in
\begin{center}
\begin{tabular}{lccccc}
\toprule
Dataset      & MUTAG        & PTC          & COX2         & PROTEINS     & NCI1         \\ \midrule
\# graphs    & 188          & 344          & 467          & 1,113        & 4,110        \\
Avg \# nodes & 17.9         & 25.5         & 41.2         & 39.1         & 29.8         \\ 
Classes      & 2            & 2            & 2            & 2            & 2            \\ 
\bottomrule
\end{tabular}
\end{center}
\vskip -0.1in
\end{table*}

\subsection{Model Architectures}
In terms of model variants, we use symmetric normalized GIND for the PPI, Chameleon and Squirrel datasets, and row-normalized GIND for the other datasets. 
We use a $4$-layer model for PPI and a $3$-layer model for the two large datasets, Chameleon and Squirrel, as well as all the datasets used for graph-level tasks. For the rest datasets, we adopt the model with only one layer. We use linear output function for all the node-level tasks, and MLP for all the graph-level tasks. 
We adopt the layer normalization (LN) \cite{ba2016layer} for all the node-level tasks and instance normalization (IN) \cite{ulyanov2016instance} for all the graph-level tasks. They compute the mean and variance used for normalization on a single training case, such that they are independent on the mini-batch size. 

\subsection{Hyperparameters}
In terms of hyperparameters, we tune learning rate, weight decay, $\alpha$ and iteration steps through the Tree-structured Parzen Estimator approach \cite{akiba2019optuna}. The hyperparameters for other baselines are consistent with those reported in their papers. Results with identical experimental settings are reused from previous works. 
\end{document}